\documentclass{article}
\usepackage{spconf,amsmath,graphicx,url}

\usepackage{booktabs}
\usepackage{amsfonts}
\usepackage[dvipsnames]{xcolor}

\title{Two Heads Better than One: Dual Degradation Representation for Blind Super-Resolution}

\name{
\begin{tabular}{@{}c@{}}
Hsuan Yuan$^{\star}$, \qquad Shao-Yu Weng$^{\star}$, \qquad I-Hsuan Lo$^{\star}$, \qquad Wei-Chen Chiu$^{\star}$, \\
Yu-Syuan Xu$^{\dagger}$, \qquad Hao-Chien Hsueh$^{\star}$, \qquad Jen-Hui Chuang$^{\star}$, \qquad Ching-Chun Huang$^{\star}$
\end{tabular}}
\address{$^{\star}$ National Yang Ming Chiao Tung University, \qquad $^{\dagger}$ MediaTek Inc. \\
        \{yuan040686.cs10, vivian5190.ee10, essen900718.cs12\}@nycu.edu.tw, walon@cs.nycu.edu.tw, \\
        yu-syuan.xu@mediatek.com, ck1040909.cs11@nycu.edu.tw, \{jchuang, chingchun\}@cs.nycu.edu.tw
}

\begin{document}

\maketitle

\newcommand{\CCH}[1]{{\color{black}#1}\normalfont} 

\begin{abstract}


\CCH{
Previous methods have demonstrated remarkable performance in single image super-resolution (SISR) tasks with known and fixed degradation (e.g., bicubic downsampling). However, when the actual degradation deviates from these assumptions, these methods may experience significant declines in performance. In this paper, we propose a Dual Branch Degradation Extractor Network to address the blind SR problem. While some blind SR methods assume noise-free degradation and others do not explicitly consider the presence of noise in the degradation model, our approach predicts two unsupervised degradation embeddings that represent blurry and noisy information. The SR network can then be adapted to blur embedding and noise embedding in distinct ways. Furthermore, we treat the degradation extractor as a regularizer to capitalize on differences between SR and HR images. Extensive experiments on several benchmarks demonstrate our method achieves SOTA performance in the blind SR problem.
}

\end{abstract}

\CCH{
\section{Introduction}
\label{sec:intro}

SISR tasks aim to reconstruct high-resolution ( HR) images from low-resolution (LR) inputs. While significant progress has been made in SISR, real-world degradation is often diverse and unknown, leading to performance drops in blind super-resolution (blind SR) settings. To tackle this issue, blind SR methods \cite{lian2023kernel,gu2019blind,luo2020unfolding,zhang2021designing,
zhang2020deep,BellKligler2019BlindSK,luo2022deep,Wang2021Unsupervised,zhou2022joint} have been proposed to handle unknown and diverse degradation.

\begin{figure}[t]
\centering

\includegraphics[width=0.46\textwidth]
{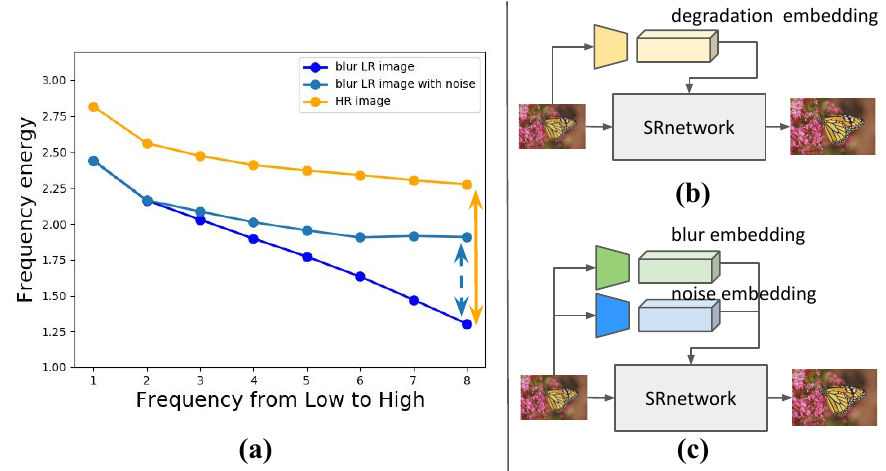}
\caption{(a) Analysis in the frequency domain among HR image, LR image, and LR image with noise. Noise degradation wrongly increases the high-frequency components, indicated by the blue arrow. Meanwhile, blur degradation leads to the loss of these high-frequency details, indicated by the orange arrow. (b) and (c) illustrate the concepts of previous UDP methods and our approach.}
\label{fig:teaser}
\end{figure}

The blind SR process involves restoring LR images based on the estimated degradation. Existing blind SR approaches can be divided into two categories based on degradation extraction designs: Supervised Kernel Prediction (SKP) and Unsupervised Degradation Prediction (UDP). SKP methods \cite{gu2019blind,luo2020unfolding,zhang2020deep,kim2021koalanet,liang2021mutual,Tao2021SpectrumtoKernelTF,jie2022DASR} primarily aim to model the signal formation process and explicitly estimate the degradation kernel, leading to visible artifacts due to minor kernel estimation errors. Besides, they mainly focus on addressing the effects of blur degradation, neglecting other types commonly encountered in real-world SR tasks. To address the drawbacks of SKP, UDP methods have been proposed to model the degradation of unknown signals implicitly. These methods employ unsupervised or self-supervised methods to estimate various degradation factors. DASR \cite{Wang2021Unsupervised} pioneers a novel technique to learn degradation embedding through contrastive learning in an unsupervised manner. Subsequently,  \cite{zhou2022joint,Wang2022DAA} expand the DASR concept and achieve improved SR outcomes. While some outperform SKP methods, they still face limitations in handling more complex degradation involving both blur and noise effects.

Existing SOTA studies \cite{Wang2021Unsupervised,zhou2022joint, jie2022DASR, Wang2022DAA} have begun to consider the noise effect on SR, but they still apply single degradation representation to account for the impact of both factors. However, as illustrated in Fig. \ref{fig:teaser}(a), blur degradation removes high-frequency content, while noise degradation introduces high-frequency pollution. That is, in the SR process, we should \textbf{recover high-frequency information} to mitigate the blur effect and \textbf{eliminate unpleasant high-frequency noise}. Therefore, we argue that using a unified degradation representation for both factors is inappropriate. Blur and noise factors should be effectively disentangled and addressed separately.


This paper introduces a method called ``Dual Degradation Representation for Blind Super-Resolution (DDSR),'' which is specifically designed to handle complex degradation. It estimates two degradation embeddings for blur and noise factors in an unsupervised manner, enabling the SR network to better adapt to diverse degradation scenarios. 
The primary contributions of this paper are outlined as follows:
\begin{itemize}
    \item \ We examine the relationship between HR images and their degraded LR counterparts, proposing a network to extract blur embeddings and noise embeddings.
\end{itemize}
\vspace{-1.2em}
\begin{itemize}
    \item \ With wavelet high-frequency inputs and learning constraints, Dual Branch Degradation Extractor produces precise degradation embeddings for blind SR.
\end{itemize}
\vspace{-1.2em}
\begin{itemize}
    \item \ Leveraging the degradation extractor as a regularizer, we introduce an additional loss function to optimize the entire network, resulting in enhanced SR images.
\end{itemize}
\vspace{-1.2em}
\begin{itemize}
    \item \ Compared to UDP methods, our approach excels in blur kernel and noise estimation. Extensive experiments on various degradation show that our method outperforms SOTA methods on blind SR.
\end{itemize}

\begin{figure*}[ht]
\centering
\includegraphics[width=\textwidth]{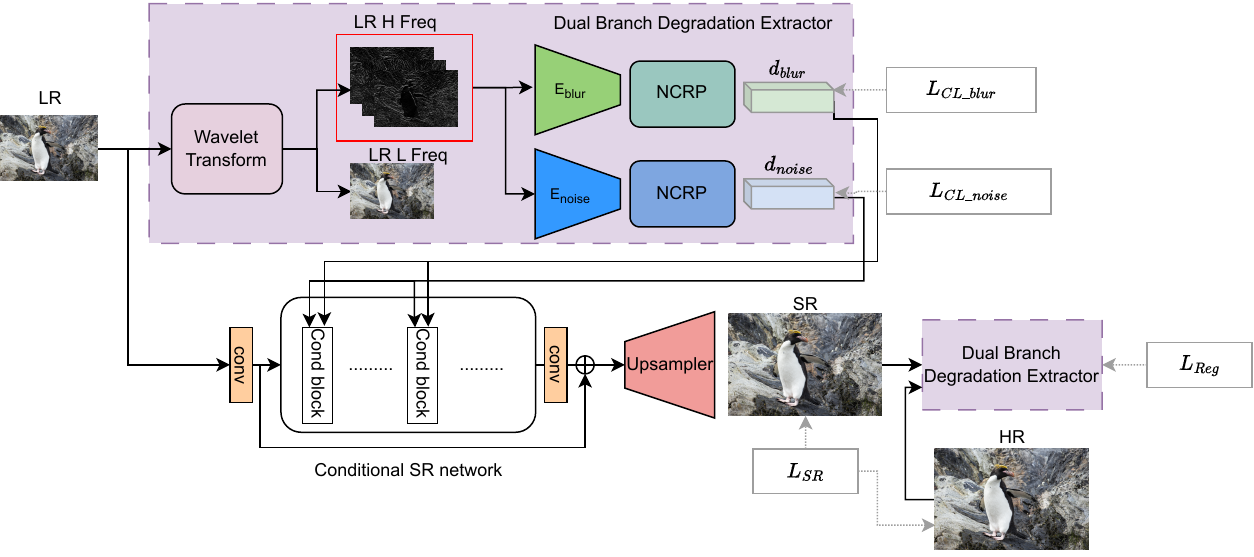}
\caption{An overview of our proposed method.}
\label{fig:network}
\end{figure*}
} 

\begin{table*}\footnotesize
\centering
  \caption{PSNR $\uparrow$ and SSIM $\uparrow$ results achieved on in-domain synthetic Urban100 x4SR. $N$ is the noise level. $S/U$ denotes SKP or UKP. The best performance is \textcolor{red}{red}, the second best is \textcolor{blue}{blue}. Upper bound is represented by underline.}
  \label{tab:syn_aniso_compare_OURx4}
  \begin{tabular}{c|cc|ccccc}
    \toprule
    \multicolumn{1}{c|}{Method} & 
    \multicolumn{1}{c}{$N$} & \multicolumn{1}{c}{$S/U$} &  
    \multicolumn{5}{c}{Blur Kernel} \\
    \multicolumn{1}{c|}{} &
    \multicolumn{2}{c|}{} &    
    $\includegraphics[scale=0.7]{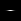}$ &
    $\includegraphics[scale=0.7]{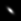}$ &
    $\includegraphics[scale=0.7]{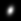}$ &
    $\includegraphics[scale=0.7]{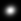}$ &
    $\includegraphics[scale=0.7]{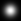}$\\    
    \midrule    DnCNN\cite{zhang2017beyond}+IKC\cite{gu2019blind} 
    & 10 &$S$
    &23.02/0.6438 &22.46/0.6126 &21.58/0.5668 
    &20.82/0.5273 &20.30/0.5017 \\
    & 20 &$S$
    &22.11/0.5920 &21.60/0.5648 &20.97/0.5335 
    &20.43/0.5070 &20.03/0.4891  \\

    DnCNN\cite{zhang2017beyond}+DCLS\cite{luo2022deep} 
    & 10 &$S$
    &23.13/0.6463 &22.43/0.6088 &21.67/0.5682 
    &21.03/0.5339 &20.54/0.5103 \\
    & 20 &$S$
    &22.30/0.5980 &21.80/0.5702 &21.21/0.5398 
    &20.68/0.5131 &20.26/0.4942   \\ 
    D-AdaptiveSR\cite{jie2022DASR}
    & 10 &$S$
    &23.21/0.6632 &22.91/0.6385 &22.19/0.5984 
    &21.42/0.5567 &20.82/0.5253  \\
    &20 &$S$
    &22.35/0.6130 &22.11/0.5926 &21.61/0.5626 
    &21.02/0.5300 &20.49/0.5032  \\     
    DASR\cite{Wang2021Unsupervised}
    & 10 &$U$
    &24.07/0.6994  &\textcolor{blue}{23.67}/0.6772 &23.13/0.6487  
    &22.59/0.6199  &21.99/0.5855 \\
    & 20 &$U$
    &23.33/0.6631 &22.93/0.6400 &\textcolor{blue}{22.41}/0.6117  &21.90/0.5840 &21.37/0.5552 \\    
    CDSR\cite{zhou2022joint}
    & 10 &$U$
    &22.07/0.6307 &22.18/0.6177 &21.46/0.5684 
    &20.92/0.5338 &20.48/0.5106  \\
    & 20 &$U$
    &21.83/0.6101 &21.86/0.5961 &21.24/0.5550 
    &20.75/0.5246 &20.34/0.5039  \\
  
    DAA\cite{jie2022DASR}
    & 10 &$U$
    &\color{blue}24.13/0.7006 &\textcolor{red}{23.77}/\textcolor{blue}{0.6801} &\color{red}23.25/0.6531 
    &\textcolor{red}{22.71}/\textcolor{blue}{0.6230} &\color{blue}22.04/0.5868  \\
    &20 &$U$
    &\color{blue}23.35/0.6648 &\color{blue}23.01/0.6438 &\color{red}22.50/0.6169
    &\color{blue}21.99/0.5877 &\color{blue}21.41/0.5568  \\ 
   
    Ours
    & 10 &$U$
    &\color{red}24.17/0.7019  &\color{red}23.77/0.6813 &\color{blue}23.22/0.6528  
    &\textcolor{blue}{22.69}/\textcolor{red}{0.6243} &\color{red}22.13/0.5919  \\

    & 20 &$U$
    &\color{red}23.41/0.6661  &\color{red}23.02/0.6446 &\textcolor{red}{22.50}/\color{blue}0.6167
    &\color{red}22.00/0.5892  &\color{red}21.49/0.5604  \\
    \hline
    Upper bound
    & 10 &
    &\underline{24.55/0.7182}
    &\underline{24.18/0.6997} 
    &\underline{23.61/0.6717} 
    &\underline{22.99/0.6399} 
    &\underline{22.43/0.6113}  \\
    & 20 &
    &\underline{23.68/0.6795} 
    &\underline{23.30/0.6596} 
    &\underline{22.78/0.6329} 
    &\underline{22.23/0.6038} 
    &\underline{21.75/0.5790}  \\    
    \bottomrule
  \end{tabular}
\end{table*}

\begin{table*}[t]\small
\caption{PSNR $\uparrow$ and SSIM $\uparrow$ results achieved on out-domain synthetic Set14 x4SR. $N$ is the noise level. $S/U$ denotes SKP or UKP. The best performance is \textcolor{red}{red}, the second best is \textcolor{blue}{blue}.}

  \label{tab:syn_aniso_out_domain}
  \begin{tabular}{c|cc|cccccc}
    \toprule
    \multicolumn{1}{c|}{Method} & 
    \multicolumn{1}{c}{$N$} & 
    \multicolumn{1}{c}{$S/U$} & 
    \multicolumn{5}{c}{Blur Kernel} \\
    \multicolumn{1}{c|}{} &
    \multicolumn{2}{c|}{} &    
    $\includegraphics[scale=0.7]{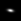}$ &
    $\includegraphics[scale=0.7]{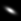}$ &
    $\includegraphics[scale=0.7]{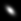}$ &
    $\includegraphics[scale=0.7]{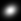}$ &
    $\includegraphics[scale=0.7]{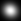}$ &
    $\includegraphics[scale=0.7]{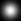}$\\    
    \midrule    
DnCNN\cite{zhang2017beyond}+IKC\cite{gu2019blind}  &
  30 &$S$&
  24.22/0.6152 &
  23.30/0.5801 & 
  23.20/0.5732 &
  22.29/0.5412 &
  21.94/0.5286 &
  21.86/0.5248 \\
  
  &40 &$S$&
  23.49/0.5941 &
  22.79/0.5640 &
  22.68/0.5588 &
  21.89/0.5305 &
  21.57/0.5189 &
  21.51/0.5163 \\

  &50 &$S$&
  22.85/0.5751 &
  22.23/0.5491 &
  22.13/0.5446 &
  21.43/0.5192 &
  21.21/0.5104 &
  21.12/0.5080 \\ 

\multicolumn{1}{c|}{DnCNN\cite{zhang2017beyond}+DCLS\cite{luo2022deep} } &
  30 &$S$&
  23.84/0.5550 &
  23.44/0.5540 &
  22.55/0.5153 &
  21.83/0.4846 &
  21.52/0.4728 &
  21.44/0.4694 \\
  
  &40 &$S$&
  22.62/0.4920 &
  22.36/0.4920 &
  21.68/0.4552 &
  21.09/0.4274 &
  20.84/0.4164 &
  20.77/0.4135 \\

  &50 &$S$&
  21.36/0.4312 &
  21.05/0.4112 &
  20.52/0.3762 &
  20.06/0.3506 &
  19.88/0.3407 &
  19.80/0.3380 \\
\multicolumn{1}{c|}{D-AdaptiveSR\cite{jie2022DASR}} &
  30 &$S$&
  22.22/0.5460 &
  21.75/0.5242 &
  21.69/0.5201 &
  21.75/0.5242 &
  20.77/0.4826 &
  20.69/0.4796 \\
  
  & 40 &$S$&
  21.43/0.4696 &
  21.01/0.4513 &
  20.96/0.4486 &
  21.01/0.4513 &
  20.22/0.4210 &
  20.16/0.4190 \\

  & 50 &$S$&
  20.56/0.3903 &
  20.22/0.3737 &
  20.18/0.3714 &
  20.21/0.3737 &
  19.60/0.3484 &
  19.55/0.3468 \\

\multicolumn{1}{c|}{DASR\cite{Wang2021Unsupervised}} &
  30 &$U$&
  24.67/0.6320 &
  23.87/0.6011 &
  23.81/0.5963 &
  22.95/0.5650 &
  \color{blue}22.45/0.5471 &
  \textcolor{red}{22.36}/\color{blue}0.5430 \\
  
  &40 &$U$&
  23.65/0.5969 &
  23.06/0.5701 &
  23.02/0.5672 &
  22.27/0.5396 &
  21.88/0.5239 &
  21.79/0.5194 \\

  & 50 &$U$&
  22.82/0.5693 &
  22.42/0.5480 &
  22.36/0.5439 &
  21.69/0.5174 &
  21.49/0.5094 &
  21.33/0.5045 \\

\multicolumn{1}{c|}{CDSR\cite{zhou2022joint}} &
  30 &$U$&
  23.55/0.5958 &
  23.55/0.5958 &
  22.86/0.5589 &
  22.07/0.5270 &
  21.77/0.5156 &
  21.68/0.5125 \\
  
 & 40 &$U$&
 21.13/0.3820 &
 20.80/0.3576 &
 20.73/0.3517  &
 20.28/0.3275  &
 20.09/0.3182  &
 20.04/0.3156  \\

 & 50 &$U$&
 20.01/0.3177  &
 19.75/0.2957  &
 19.69/0.2904  &
 19.33/0.2684  &
 19.17/0.2598  &
 19.13/0.2574  \\

\multicolumn{1}{c|}{DAA\cite{Wang2022DAA}} &
  30 &$U$&
  \color{red}24.73/0.6335 &
  \color{blue}23.95/0.6041 &
  \textcolor{red}{23.89}/\color{blue}0.5995 &
  \color{blue}22.96/0.5663 &
  \color{red}22.46/0.5483 &
  \color{red}22.36/0.5441 \\
  
  & 40 &$U$&
  \color{blue}23.93/0.6061 &
  \color{blue}23.33/0.5814 &
  \color{blue}23.22/0.5760 &
  \color{blue}22.42/0.5475 &
  \color{blue}22.00/0.5319 &
  \color{blue}21.91/0.5278 \\

  & 50 &$U$&
  \color{blue}23.16/0.5719 &
  \color{blue}22.62/0.5491 &
  \color{blue}22.54/0.5452 &
  \color{blue}21.79/0.5196 &
  \color{blue}21.56/0.5117 &
  \color{blue}21.42/0.5058 \\

\multicolumn{1}{c|}{Ours} 

  &30 &$U$&
  \color{blue}24.70/0.6322 &
  \color{red}23.97/0.6053 &
  \textcolor{blue}{23.87}/\color{red}0.6000 &
  \color{blue}22.98/0.5664 &
  \textcolor{red}{22.46}/0.5464 &
  \textcolor{blue}{22.31}/0.5408 \\

  & 40 &$U$&
  \color{red}24.08/0.6119 &
  \color{red}23.43/0.5875 &
  \color{red}23.34/0.5828 &
  \color{red}22.52/0.5520 &
  \color{red}22.09/ 0.5358 &
  \color{red}21.97/0.5315 \\

  & 50 &$U$&
  \color{red}23.55/0.5951 &
  \color{red}22.95/0.5721 &
  \color{red}22.87/0.5678 &
  \color{red}22.13/0.5403 &
  \color{red}21.77/0.5272 &
  \color{red}21.67/0.5237 \\

  \bottomrule
\end{tabular}
\end{table*}

\CCH{

\begin{figure}
\centering
\includegraphics[width=0.45\textwidth]{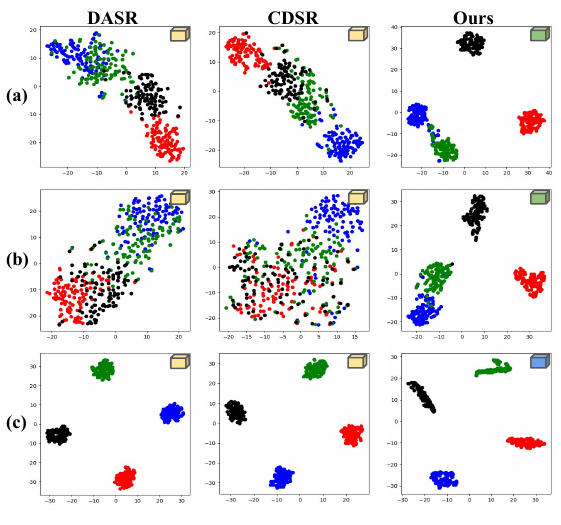}
\caption{Visualization of different degradations using colored dots in three settings. (a) w/
different blur kernels and w/o noise level. (b) w/ different blur kernels and noise level is 10. (c) w/ a fixed kernel and different noise levels. \textcolor{yellow}{Yellow} box means using a unified degradation representation, \textcolor{green}{green} box denotes blur representations, and \textcolor{blue}{blue} box indicates noise representations. Compared with DASR and CDSR, our method can clearly distinguish different degradations in all the test settings (i.e., (a), (b), and (c)). }
\label{fig:tnse_compare_with_other}
\end{figure}

\begin{figure*}[ht]
\centering
\includegraphics[width=1\textwidth]
{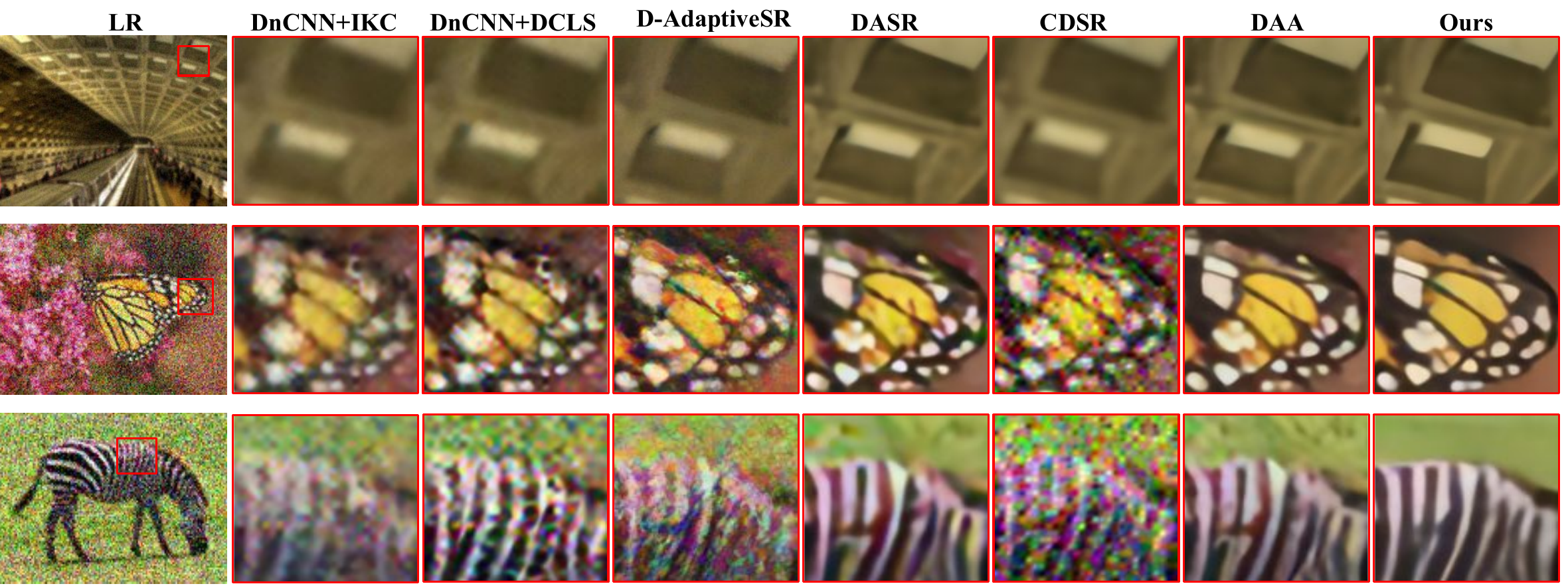}
\caption{Visual comparison on in-domain (first row) and out-domain (second and thrid row)  synthetic test data for $\times4 $ SR.}
\label{fig:visual_compare}
\end{figure*}

\vspace{-0.5em}
\section{Proposed Method}
\subsection{Problem Formulation}\label{Problem Formulation} 

The image degradation process is formulated as follows:
\begin{equation}
I^{LR}=fd_k\left ( I^{HR}, k\right )\downarrow_{s}+n,
\label{eq.lr synthesize}
\end{equation}  
where $I^{LR}$ denotes the LR image, $I^{HR}$ indicates the HR image, $k$ is a blur kernel, $fd_k$ represents the blind blur function, $\downarrow_{s}$ is downsampling operator with scale factor $s$, and $n$ refers to additive white Gaussian noise. In the blind SR task, the objective is to restore HR images from LR counterparts, given unknown blur kernels and noise levels. 

\subsection{Dual Branch Degradation Extractor}\label{Dual Branch degradation extractor}
Our Dual Branch Degradation Extractor, shown in Fig. \ref{fig:network}, is designed to extract and disentangle blur and noise factors from LR input images. The network comprises two branches working collaboratively to extract blur and noise representations. These representations later assist our SR network in effectively producing high-quality, high-resolution outputs. The following subsections delve into the three primary modules constituting our degradation extractor.

\noindent\textbf{(a) Degradation Extraction in the Wavelet Domain:}
Contrary to previous methods \cite{Wang2021Unsupervised,zhou2022joint} directly process LR images, our DDSR employs wavelet transform to analyze high-frequency components. The high-frequency components, housing fine details, and textures, are susceptible to noise and blur degradation. Focusing on the wavelet domain, specifically high-frequency components, makes our extractor more flexible in identifying the differences between blur and noise factors. The separation leads to more accurate degradation representations, improving super-resolution performance.

\noindent\textbf{(b) Constraints on learning blur and noise representations:}
To ensure accurate embedding learning, we implement techniques to constrain them to contain relevant information while minimizing irrelevant information by contrastive learning with positive and negative samples. Given a query image $I^{query}$, degraded by blur kernel $k$ and noise level $n$, we generated its blur embedding $d_{k}$ and noise embedding $d_{n}$ through the blur and noise extractors, respectively.
The positive ``blur'' embedding $d^{+}_{k}$ is acquired from $I^{+}_{blur}$ using the blur extractor, where $I^{+}_{blur}$ is degraded by the same blur kernel $k$ like $I^{query}$ but a different noise level ${n}'$. By contrast, the negative ``blur'' sample $d^{j-}_{k}$ is extracted from another training image $I^{j-}_{blur}$, which is degraded by various other blur kernels and noise levels. Similarly, the positive ``noise'' embedding $d^{+}_{n}$ is obtained from $I^{+}_{noise}$ that has been degraded by a different blur kernel ${k}'$ but the same noise level $n$, while the negative sample $d^{j-}_{n}$ is extracted from $I^{j-}_{noise}$ subjected to different blur kernels and noise levels. In brief, to train the blur extractor, the query and positive samples are degraded by the same kernel but with different noise levels, while negative samples in the queue undergo degradation by other kernels and noise levels. A similar approach is followed to train the noise extractor, ensuring precise learning of degradation embeddings for effective discrimination.

\noindent\textbf{(c) Normalized Codebook-based Representation Purification:}
We introduce Normalized Codebook-based Representation Purification (NCRP). During training, we learn basis feature vectors to extract a compact noise representation space (i.e., defined by another $L$ kernel basis vectors $\{cb^j_{k}\}_{j=1:L}$). As the extracted low-dimensional spaces contain pure degradation information, we can project the extracted raw noise and blur embeddings $\{q_{n},q_{k}\}$ onto these spaces to remove irrelevant redundancies and achieve the purified noise and kernel representations $\{d_{n},d_{k}\}$. To realize the idea, we use a codebook-based spatial compression module, expressed mathematically as:
\begin{equation}
\begin{aligned}
&q_{i}=Norm\left ( \phi \left ( E_{i}\left ( \omega \left ( I^{LR} \right ) \right ) \right ) \right ),\\ 
& k^j_{i}=Norm\left ( \phi \left ( cb^j_{i} \right ) \right ),\\
& K_{i}=\left [ k^{1}_{i},k^{2}_{i},...,k^{L}_{i} \right ]^{T},~and\\
&d_{i}=Softmax\left (  q_{i}\cdot K_{i}^{T} \right )\cdot CB_{i}.
\end{aligned}
\label{eq.codebook}
\end{equation}
\noindent
where $\omega$ is the wavelet transform operator for extracting high-frequency components and $\omega \left ( I^{LR} \right )$ denotes the input of two degradation extractors. Here, we use $\{E_{i}\}$ to denote the noise and kernel extractors, where $i\in\{n, k\}$. Furthermore, $\phi$ represents an MLP projection, $Norm$ denotes vector normalization, and $CB_{i}\in \mathbb{R^{L\times N}}$ are the learned codebooks for $i\in\{n, k\}$. The codebook size is $L$, and the embedding dimension is $N$. Adding a normalization layer in our codebook-based spatial compression module enhances the training process stability, as observed in our experiments.

\subsection{Conditional SR network}\label{SRnetwork}
After obtaining the blur and noise embeddings from the Dual Branch Degradation Extractor, our SR network uses these embeddings to restore each LR image adaptively. The SR scheme is flexible, as the backbone of the SR network is replaceable. However, modifications are still necessary to accommodate the two degradation representations as extra conditions. In Fig. \ref{fig:network}, ``Cond Block'' serves as the conditional block to insert the blur and noise representations into the SR network. We have modified the SR network architectures used in DASR\cite{Wang2021Unsupervised} to implement the ``Cond Block''. Here, we realized an adaptive SR network to integrate information from the blur degradation embedding with DASR and followed a similar approach to introduce the other information channel for noise degradation embedding by adding a side network. The side network stretches the noise embedding to match the size of the feature map, where we insert the conditional embedding within the SR network. Next, the concatenated noise embeddings and feature maps are fused and inputted into the SR network through a convolutional layer. This enables the SR network to better account for the noise effect for different input LR images and dynamically improve the restoration results. Due to limited space, we provide the ``Cond Block'' details, our modified SR network, and more discussions \textcolor{black}{in the supplementary~\cite{ddsr}. Codes are available upon acceptance.}

\subsection{Total Loss Functions}\label{Total Loss Functions}
\noindent\textbf{Contrastive Loss $L_{CL}$:} For training Dual Branch Degradation Extractor, we utilize the InfoNCE contrastive learning loss, defined as:
\begin{equation} L_{CL\_i}=\sum_{m=1}^{B} -log\left ( \frac{exp\left ( d^{m}_{i}\cdot d^{m+}_{i}/\tau\right )}{\sum_{j=1}^{Nqueue} \left ( exp(d^{m}_{i}\cdot d^{j-}_{i}/\tau) \right )} \right ). \label{eq.contrative} \end{equation}
\noindent Here, $i\in\{n, k\}$ indicates the noise or blur degradation, $m$ is the index of a query sample, $d^{m+}_{i}$ is the positive key embedding corresponding to the query embedding $d^m_{i}$; $Nqueue$ is the number of negative samples in the queue, $d^{j-}_{i}$ denotes the $j$-th negative sample, and $\tau$ is the temperature hyper-parameter. For every batch, we use $B$ LR images for training. As mentioned in Sec.\ref{Dual Branch degradation extractor}, queries and positive keys crop from different images further polish the degradation extractor.

\noindent\textbf{SR Restoration Loss $L_{SR}$:} We employ standard restoration loss $L_{SR}$ to train the SR network, measuring $L_1$ difference between super-resolved (SR) images and HR images. This helps the SR network to produce images that are visually similar to ground truth, achieving more accurate SR results.

\noindent\textbf{Regularization Loss $L_{Reg}$:}
We introduce $L_{Reg}$ loss to compare the degradation embeddings of predicted SR images with HR embeddings obtained from the same extractor (either noise or blur extractor). By minimizing the embedding difference using $L_1$ distance, the network is guided to generate an SR output that closely aligns with the HR image, thereby improving the quality of the SR results.

The total loss function can be formulated as follows:
\begin{equation}
\begin{aligned}
&L_{total} = L_{CL} + L_{SR} + L_{Reg}\\
&L_{CL} = L_{CL\_noise} + L_{CL\_blur}\\
&L_{Reg} = \lambda_{1}\cdot L_{Reg\_noise} + \lambda_{2}\cdot L_{Reg\_blur},\\
\end{aligned} 
\end{equation}
where $\lambda_{1}$ is 1000 and $\lambda_{2} $ is 10. 
}

\begin{figure}
\centering
\includegraphics[width=0.45\textwidth]{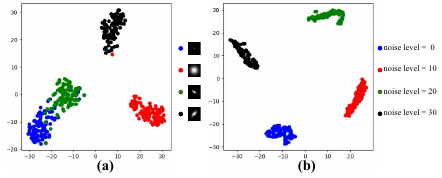}
\caption{(a) Distribution of \textbf{``blur representations''} with different blur kernels, each sample has a random noise level. (b)  Distribution of \textbf{``noise representations''} with different noise levels, each sample comes from a random blur kernel.}
\label{fig:tsne_ours}
\end{figure}

\CCH{

\begin{figure}
\centering
\includegraphics[width=0.47\textwidth]{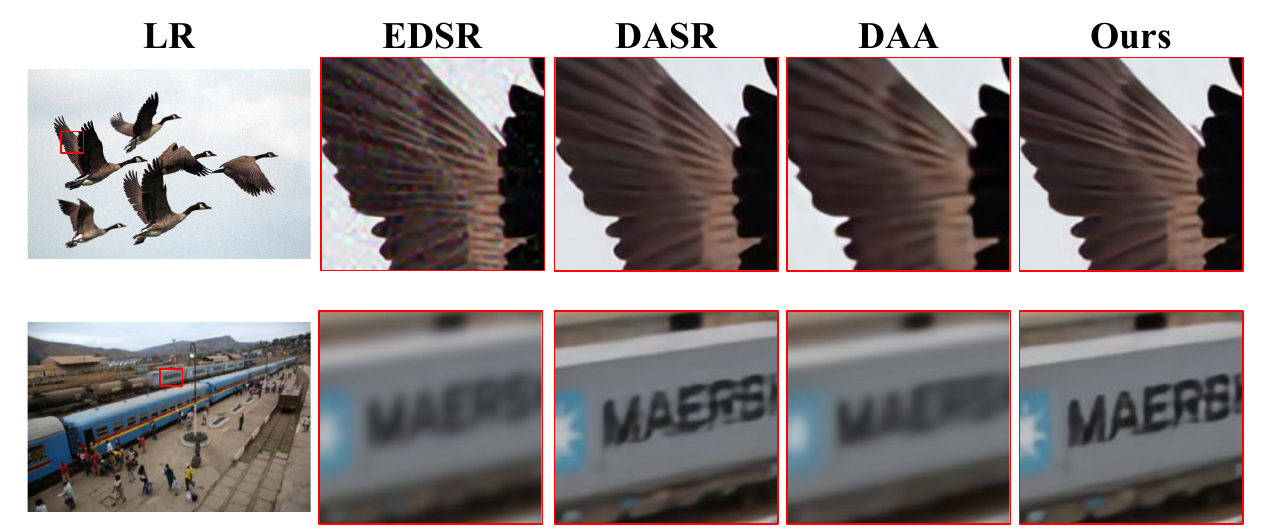}
\caption{Qualitative comparison on NTIRE2020Track1 (first row) and DIV2KRK (second row).}
\label{fig:generalization}
\end{figure}

\section{Experiments}
\subsection{Datasets and Implementation Details}
\noindent\textbf{Implementation Details.} 
We synthesize LR images by the degradation process formulated in Eq.(\ref{eq.lr synthesize}). Training set consists of 800 images from DIV2K \cite{8014884} and 2650 images from Flickr2K \cite{8014883}. Anisotropic Gaussian kernels are used for degradation with kernel size fixed to 21$\times$21 and kernel widths $\lambda_{1}$, $\lambda_{2}$ uniformly sampled from the range [0.2,4] for $\times$4 downsample. The range of rotation angle of Gaussian kernels is  [0, $\pi$]. Noise levels range from [0, 25]. The LR image patch size is set to 48$\times$48, and the batch size to 32.

While training Dual Branch Degradation Extractor, we optimize $L_{cl}$ with a learning rate of $1 \times 10^{-3}$. Then, fixing the extractor and training SR framework with regularizers, minimizing $L_{SR}+L_{Reg}$ with a learning rate of $1 \times 10^{-4}$. Finally, we fine-tune the entire network with regularizers to optimize the total loss. $\tau$ and $N_{queue}$ in Eq.(\ref{eq.contrative}) are set to  0.07 and 8192, respectively. Codebook size $L$ is set to 128, and embedding dimension $N$ is set to 64. Adam optimizer with $\beta_{1}=0.9$ and $\beta_{2}=0.999$ is used for optimization.

\noindent\textbf{Performance Comparison.} 
Four benchmark datasets, Set5 \cite{Bevilacqua2012LowComplexitySS}, Set14 \cite{Zeyde2010OnSI}, B100 \cite{Martin2001ADO}, and Urban100 \cite{7299156}, are used for blind SR evaluation. Trained models are directly tested on DIV2KRK \cite{BellKligler2019BlindSK}, DIV2KRK w/ noise level 10, and NTIRE 2020Track1 \cite{9150677} without fine-tuning. The performance of the models is assessed using PSNR and SSIM metrics, calculated on the Y channel of the SR images in the YCbCr space. 

\vspace{-0.5em}
\subsection{Analysis on Degradation Representation}
We perform t-SNE analysis of DASR, CDSR, and our method for comparison, as these methods share similar assumptions about degradation representations. Following \cite{Wang2021Unsupervised}, we use the B100 dataset to generate LR images with different degradations and then feed them to the degradation extractor of each method to obtain corresponding degradation representations. As shown in Fig. \ref{fig:tnse_compare_with_other}, our method produces more separable clusters between blur types. Under the influence of noise, our process outperforms DASR and CDSR and remains effective in distinguishing blur kernels. More robustness tests are shown in Fig. \ref{fig:tsne_ours}, where LR images with diverse degradations are used to evaluate our blur and noise degradation extractors. Results indicate that our embeddings can effectively handle various and mixed degradations.


\vspace{-0.5em}
\subsection{Experiments on General Degradations}
For evaluating degradations under diverse anisotropic Gaussian kernels and noise levels, our method is compared to baseline and SOTA blind SR models with 11 typical blur kernels and different noise levels. In-domain data comprises degradations observed during training, and out-domain data does not. For fair comparisons, we retrain related SOTA methods under the same experimental setting. To handle various noise levels, we denoise LR images using DnCNN \cite{zhang2017beyond} under a blind setting for SKP methods \cite{gu2019blind,luo2022deep}. Results in Tab. \ref{tab:syn_aniso_compare_OURx4}, Tab. \ref{tab:syn_aniso_out_domain}, and Fig. \ref{fig:visual_compare} show our comparable performace on in-domain synthesic Urban100\cite{7299156} and outperformance on out-domain synthesic Set14\cite{Zeyde2010OnSI}. Additionally, we investigate the upper bound of our method by providing GT blur kernels and noise levels, shown in Tab. \ref{tab:syn_aniso_compare_OURx4}. The results indicate that our method is close to the upper bound.
From the results, DnCNN\cite{zhang2017beyond}+IKC\cite{gu2019blind} and DnCNN\cite{zhang2017beyond}+DCLS\cite{luo2022deep} output the artificial SR results as DnCNN \cite{zhang2017beyond} removes noise but distort images simultaneously. Compared with other UDP methods, our method can provide better qualitative and quantitative results due to proper degradation estimations and suitable handling of different degradation representations separately.

\vspace{-0.5em}
\subsection{Model Generalizability}
\label{sec.generalization}
To demonstrate generalizability, we evaluate our model on DIV2KRK, DIV2KRK w/ noise level 10, and NTIRE2020Track1. Since our method aims to handle more generalized and mixed degradations, we add noise with noise level 10 to the noise-free DIV2KRK dataset, creating a new dataset ``DIV2KRK w/ n 10'' for performance evaluation under different noise levels. Quantitative results in Tab. \ref{tab:generalization} and visualization in Fig. \ref{fig:generalization} demonstrate our model generalization across different datasets and robustness against various degradation types.
\begin{table}[t]\footnotesize
\caption{Quantitative comparison among different methods in terms of model generalization. NTIRE20T1 represents NTIRE2020Track1 dataset. The best performance is highlighted in \textcolor{red}{red}, and the second best is shown in \textcolor{blue}{blue}.}
\label{tab:generalization}
\begin{tabular}{c|ccc}
\toprule
Methods  & \multicolumn{1}{c}{DIV2KRK} & \multicolumn{1}{c}{DIV2KRK w/ n10} & \multicolumn{1}{c}{NTIRE20T1} \\
\midrule
EDSR\cite{Lim_2017_CVPR_Workshops}   & 25.64/0.6928 &21.60/0.5101  &25.34/0.6391 \\
DASR\cite{Wang2021Unsupervised}  &\color{blue}26.62/0.7682 &\color{blue}24.86/0.7063 &\color{blue} 27.21/0.7901 \\
DAA\cite{Wang2022DAA}  &24.62/0.6988 &24.10/0.6836 &26.13/0.7596\\
Ours  & \color{red}26.79/0.7745 &\color{red}24.97/0.7100 &\color{red}27.28/0.7921 \\
\bottomrule
\end{tabular}
\end{table}

\begin{table}\footnotesize
  \caption{Ablation Study on Set14 x4SR. }
  \label{tab:ablation_study}
  \begin{tabular}{c|cccc|c}
    \toprule
    \multicolumn{1}{c|}{Method} &     
    \multicolumn{1}{c}{W Trans} &
    \multicolumn{1}{c}{CC} &
    \multicolumn{1}{c}{NCRP} &
    \multicolumn{1}{c|}{$L_{Reg}$} &
    \multicolumn{1}{c}{PSNR/SSIM} \\
    \midrule
    Model1
    &$\times$ &\checkmark &\checkmark  &\checkmark 
    &25.54/0.6608   \\
    Model2
    &\checkmark  &$\times$ &\checkmark  &\checkmark 
    &25.35/0.6525  \\     
    Model3 
    &\checkmark   &\checkmark &$\times$ &\checkmark 
    &25.25/0.6496  \\
    Model4 
    &\checkmark   &\checkmark  &\checkmark &$\times$ 
    &25.46/0.6561  \\
    Model5
    &\checkmark  &\checkmark  &\checkmark &\checkmark 
    &25.60/0.6620  \\

    \bottomrule
  \end{tabular}
\end{table}
\subsection{Ablation Study}
Ablation studies reveal the significance of components in our method. As shown in Tab. \ref{tab:ablation_study}, the result of ``Model 1'' shows that using RGB images without wavelet transform (``W Trans'') as input makes the extractor training harder. Without constraints of learning blur and noise embeddings (``CC''), the network learns incorrect blur or noise embeddings, leading to significant drops in PSNR. Normalized codebook-based representation purification (``NCRP'') and using degradation extractors for model regularization (``$L_{Reg}$'') also bring significant improvements. }

\vspace{-0.5em}

\section{Conclusions}
We introduce DDSR to super-resolve images with complex image degradation. Dual Branch Degradation Extractor employs two branches to collaboratively extract blur and noise representations for guiding conditional SR network to enhance image restoration. Furthermore, degradation is used as a regularizer to minimize differences between SR and HR, leading to improved performance. Experimental result shows that DDSR achieves SOTA performance in blind SR tasks.

\clearpage
\bibliographystyle{IEEEbib}
\bibliography{main}
\end{document}